\documentclass{article}

\usepackage[preprint]{neurips_2019}

\usepackage[utf8]{inputenc} % allow utf-8 input
\usepackage[T1]{fontenc}    % use 8-bit T1 fonts
\usepackage{hyperref}       % hyperlinks
\usepackage{url}            % simple URL typesetting
\usepackage{booktabs}       % professional-quality tables
\usepackage{amsfonts}       % blackboard math symbols
\usepackage{nicefrac}       % compact symbols for 1/2, etc.
\usepackage{microtype}      % microtypography

\usepackage{graphicx}
\usepackage[ruled,vlined]{algorithm2e}
\usepackage{booktabs}
\usepackage{amsmath}
\usepackage{float}

\usepackage{natbib}
\setcitestyle{authoryear,open={(},close={)}}
\hypersetup{colorlinks,linkcolor={blue},citecolor={blue},urlcolor={red}}

\title{DMSANet: Dual Multi Scale Attention Network}

\author{%
  Abhinav Sagar\thanks{Website of author - \url{https://abhinavsagar.github.io/}} \\
  Vellore Institute of Technology\\
  Vellore, India\\
  \texttt{abhinavsagar4@gmail.com} \\

}

\begin{document}

\nocite{*}

\maketitle

\begin{abstract}
Attention mechanism of late has been quite popular in the computer vision community. A lot of work has been done to improve the performance of the network, although almost always it results in increased computational complexity. In this paper, we propose a new attention module that not only achieves the best performance but also has lesser parameters compared to most existing models. Our attention module can easily be integrated with other convolutional neural networks because of its lightweight nature. The proposed network named Dual Multi Scale Attention Network (DMSANet) is comprised of two parts: the first part is used to extract features at various scales and aggregate them, the second part uses spatial and channel attention modules in parallel to adaptively integrate local features with their global dependencies. We benchmark our network performance for Image Classification on ImageNet dataset, Object Detection and Instance Segmentation both on MS COCO dataset.
\end{abstract}

\section{Introduction}

The local receptive field of the human eye has led to the construction of convolutional neural networks which has powered much of the recent advances in computer vision. Multi scale architecture used in the famous InceptionNet \citep{szegedy2016rethinking} aggregates multi-scale information from different size convolutional kernels. Attention Networks has attracted a lot of attention recently as it allows the network to focus on only then essential aspects while ignoring the ones which are not useful \citep{li2019selective}, \citep{cao2019gcnet} and \citep{li2019selective}. 

A lot of problems have been successfully tackled using attention mechanism in computer vision like image classification, image segmentation, object detection and image generation. Most of the attention mechanisms can be broadly classified into two types channel attention and spatial attention, both of which strengthens the original features by aggregating the same feature from all the positions with different aggregation strategies, transformations, and strengthening functions \citep{zhang2021epsanet}. 

Some of the work combined both these mechanism together and achieved better results \citep{cao2019gcnet} and \citep{woo2018cbam}. The computational burden was reduced by \citep{wang2020ecanet} using efficient channel attention and $1\times1$ convolution. The most popular attention mechanism is the Squeeze-and Excitation module \citep{hu2018squeeze}, which can significantly improve the performance with a considerably low cost. The “channel shuffle” operator is used \citep{zhang2021sa} to enable information communication between the two branches. It uses a grouping strategy, which divides the input feature map into groups along the channel dimension. 

\section{Related Work}

There are two main problems which hinders the progress in this field: 1) Both spatial and channel attention as well as network using combination of two uses only local information while ignoring long range channel dependency, 2) The previous architectures fail to capture spatial information at different scales to be more robust and handle more complex problems. These two challenges were tackled by \citep{duta2020pyramidal} and \citep{li2019selective} respectivly. The problem with these architectures is that the number of parameters increased considerably. 

Pyramid Split Attention (PSA) \citep{zhang2021epsanet} has the ability to process the input tensor at multiple scales. A multi-scale pyramid convolution structure is used to
integrate information at different scales on each channel-wise feature map. The channel-wise attention weight of the multi-scale feature maps are extracted hence long range channel dependency is done.

Non-Local block \citep{wang2018non} is proposed to
build a dense spatial feature map and capture the long-range dependency using non-local operations. \citep{li2019selective}
used a dynamic selection attention mechanism that allows each neuron to adaptively adjust its receptive field size based on multiple scales of input feature map. \citep{fu2019dual} proposed a network to integrate local features with their global dependencies by summing these two
attention modules from different branches. 

Multi scale architectures have been used sucessfully for a lot of vision problems \citep{sagar2020monocular}, \citep{hu2018squeeze} and \citep{sagar2020semantic}. \citep{fu2019dual} adaptively integrated local
features with their global dependencies by summing the two
attention modules from different branches. \citep{hu2018gather} used spatial extension using a depth-wise convolution to aggregate individual features. Our network borrows ideas from \citep{gao2018channelnets} which used a network to capture local cross-channel interactions.

The performance (in terms of accuracy) vs computational complexity (in terms of number of parameters) of the state of art attention modules is shown in Figure 1:

\begin{figure}[htp]
    \centering
    \includegraphics[width=10cm]{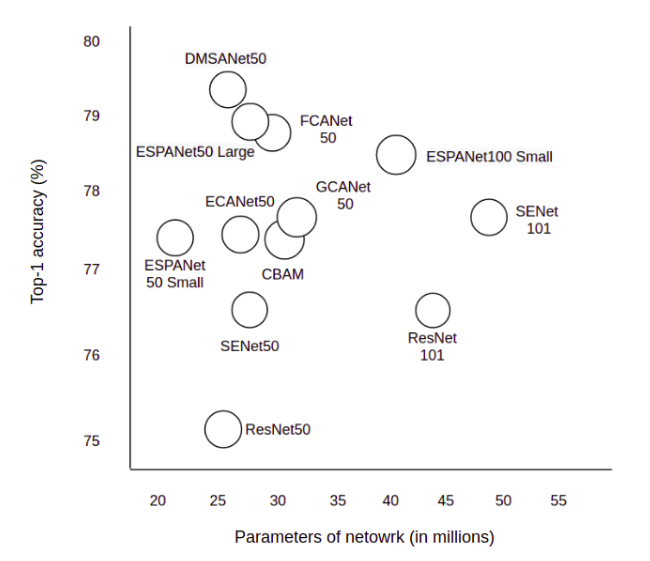}
    \caption{Comparing the accuracy of different attention methods with ResNet-50 and ResNet-101 as backbone in terms of accuracy and network parameters. The circles reflects the network parameters and FLOPs of different models. Our proposed network achieves higher
accuracy while having less model complexity.}
    \label{a1}
\end{figure}

Our main contributions can be summarized as follows:

• A new attention module is proposed which aggregates feature information at various scales. Our network is scalable and can be easily plugged into various computer vision problems.

• Our network captures more contextual information using both spatial and channel attention at various scales.

• Our experiments demonstrate that our network outperforms previous state of the art with lesser computational cost.

\section{Method}

\subsection{Feature Grouping}

Shuffle Attention module divides the input feature map into
groups and uses Shuffle Unit to integrate the channel attention and spatial attention into one block for each group. The sub-features are aggregated and a “channel shuffle” operator is used for communicating the information between different sub-features.

For a given feature map $X\in R^{C\times H\times W}$ , where $C, H, W$ indicate the channel number, spatial height, and width, respectively, shuffle attention module divides $X$ into $G$ groups along the channel dimension, i.e., $X = [X1, X_{G}], X_{k}\in R^{C/G\times H\times W}$. An attention module is used to weight the importance of each feature. The input of $X_{k}$ is split into two networks along the channel dimension $X_{k1}, X_{k2}\in R^{C/2G\times H\times W}$. The first branch is used to produce a channel attention
map by using the relationship of channels, while
the second branch is used to generate a spatial attention map by using the spatial relationship of different features.

\subsection{Channel Attention Module}

The channel attention module is used to selectively weight the importance of each channel and thus produces best output features. This helps in reducing the number of parameters of the network. Let $X\in R^{C\times H\times W}$ denotes the input feature map,
where the quantity $H, W, C$ represent its height, width and number of input channels respectively. A SE block consists of two parts: squeeze and excitation, which are respectively designed for encoding the global information and
adaptively recalibrating the channel-wise relationship. The Global Average Pooling (GAP) operation can be calculated by the as shown in Equation 1:

\begin{equation}
GAP_{c}=\frac{1}{H \times W} \sum_{i=1}^{H} \sum_{j=1}^{W} x_{c}(i, j)
\end{equation}

The attention weight of the $c^{th}$ channel in the SE block can be written as denoted in Equation 2:

\begin{equation}
w_{c}=\sigma\left(W_{1} ReLU\left(W_{0}\left(GAP_{c}\right)\right)\right)
\end{equation}

where $W_{0}\in R^{C\times C} r$ and $W_{1}\in R^{C r\times C}$ represent the fully-connected (FC) layers. The symbol $\sigma$ represents the excitation function where Sigmoid function is usually used.

We calculate the channel attention map $X\in R^{C\times C}$
from the original features $A\in R^{C\times H\times W}$. We reshape A to $R^{C\times N}$ , and then perform a matrix multiplication between A and the transpose of A. We then apply a softmax layer to obtain the channel attention map $X \in R^{C\times C}$ as shown in Equation 3:

\begin{equation}
x_{j i}=\frac{\exp \left(A_{i} \cdot A_{j}\right)}{\sum_{i=1}^{C} \exp \left(A_{i} \cdot A_{j}\right)}
\end{equation}

where $x_{ji}$ measures the $i^{th}$ channel’s impact on the $j^{th}$ channel. We perform a matrix multiplication
between the transpose of $X$ and $A$ and reshape their result
to $R^{C\times H\times W}$ . We also multiply the result by a scale parameter $\beta$ and perform an element-wise sum operation with $A$ to obtain the final output $E\in R^{C\times H\times W}$ as shown in Equation 4:

\begin{equation}
E_{1j}=\beta \sum_{i=1}^{C}\left(x_{j i} A_{i}\right)+A_{j}
\end{equation}

\subsection{Spatial Attention Module}

We use Instance Normalization (IN) over $X_{k2}$ to obtain spatial-wise statistics. A $F{c}(·)$ operation is used to enhance the representation of $X_{k2}$. The final output of spatial attention is obtained by where $W_{2}$ and $b_{2}$ are parameters with shape $R^{C/2G\times1\times1}$. After that the two branches are concatenated to make the number of channels equal to the number of input.

A local feature denoted by $A \in R^{C\times H\times W}$ is fed into a convolution layer to generate two new feature maps $B$ and $C$, respectively where ${B, C}\in R^{C\times H\times W}$. We reshape them to $R^{C\times N}$, where $N=H\times W$ is the number of pixels. Next a matrix multiplication is done between the transpose of $C$ and $B$, and apply a softmax layer to calculate the spatial
attention map $S\in R^{N\times N}$. This operation is shown in Equation 1:

\begin{equation}
s_{j i}=\frac{\exp \left(B_{i} \cdot C_{j}\right)}{\sum_{i=1}^{N} \exp \left(B_{i} \cdot C_{j}\right)}
\end{equation}

where $s_{ji}$ measures the $i^{th}$ position’s impact on $j^{th}$ position. Next we feed feature A into a convolution layer to generate a new feature map $D\in R^{C\times H\times W}$ and reshape
it to $R^{C\times N}$. We perform a matrix multiplication between $D$ and the transpose of $S$ and reshape the result to $R^{C\times H\times W}$ . We multiply it by a scale parameter $\alpha$
and perform a element-wise sum operation with the feature
$A$ to obtain the final output $E\in R^{C\times H\times W}$ as shown in Equation 2:

\begin{equation}
E_{2j}=\alpha \sum_{i=1}^{N}\left(s_{j i} D_{i}\right)+A_{j}
\end{equation}

\subsection{Aggregation}

In the final part of the network, all the sub-features are aggregated. We use a ”channel shuffle” operator to enable cross-group information flow along the channel dimension. The final output of our module is the same size as that of input, making our attention module quite easy to integrate with other networks.

The whole multi-scale pre-processed feature map can be obtained by a concatenation way as defined in Equation 7:

\begin{equation}
F=\operatorname{Concat}\left(\left[E_{1j}, E_{2j}\right]\right)
\end{equation}

where $F\in R^{C\times H\times W}$ is the obtained multi-scale feature map. Our attention module is used across channels to adaptively select different spatial scales which is guided by the feature descriptor. This operation is defined in Equation 8:

\begin{equation}
a t t_{i}=\operatorname{Softmax}\left(Z_{i}\right)=\frac{\exp \left(Z_{i}\right)}{\sum_{i=0}^{S-1} \exp \left(Z_{i}\right)}
\end{equation}

Finally we multiply the re-calibrated weight of multi-scale channel attention $a_{tti}$ with the feature map of the corresponding scale $F_{i}$ as shown in Equation 9:

\begin{equation}
Y_{i}=F_{i} \odot a t t_{i} \quad i=1,2,3, \cdots S-1
\end{equation}

\subsection{Network Architecture}

We propose DMSA module with the goal to build more efficient and scalable architecture. The first part of our network borrows ideas from \citep{li2019selective} and \citep{zhang2021sa}. An input feature map $X$ is splitted into $N$ parts along with the channel dimension. For each splitted parts, it has $C_{0}$=$C_{S}$ number of common channels, and the $i^{th}$ feature map is $X_{i}\in R^{C_{0}\times H\times W}$. The individual features are fused before being passed to two different branches.

These two branches are comprised of position attention module and channel attention module as proposed in \citep{fu2019dual} for semantic segmentation. The second part of our network does the following 1) Builds a spatial attention matrix which models the spatial relationship between
any two pixels of the features, 2) A matrix
multiplication between the attention matrix and the original
features. 3) An element-wise sum operation is done on the resulting matrix and original
features. 

The operators concat and sum are used to reshape the features. The features from the two parallel branches are aggregated to produce the final output. The complete network architecture is shown in Figure 2:

\begin{figure}[htp]
    \centering
    \includegraphics[width=10cm]{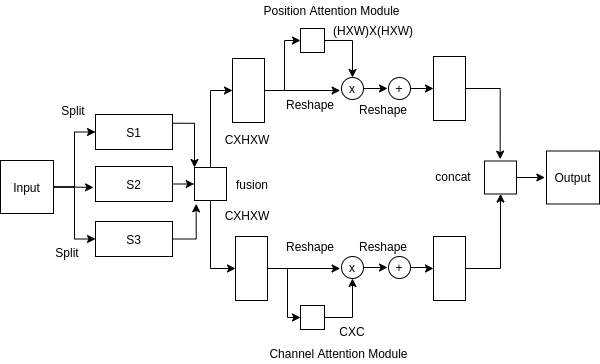}
    \caption{A detailed Illustration of DMSANet}.
    \label{a1}
\end{figure}

We compare our network architecture with Resnet \citep{wang2017residual}, SENet \citep{hu2018squeeze} and EPSANet \citep{zhang2021epsanet} in Figure 3. We use our DMSA module in between $3\times3$ convolution and $1\times1$ convolution. Our network is able to extract features at various scales and aggregate those individual features before passing through the attention module. 

\begin{figure}[htp]
    \centering
    \includegraphics[width=10cm]{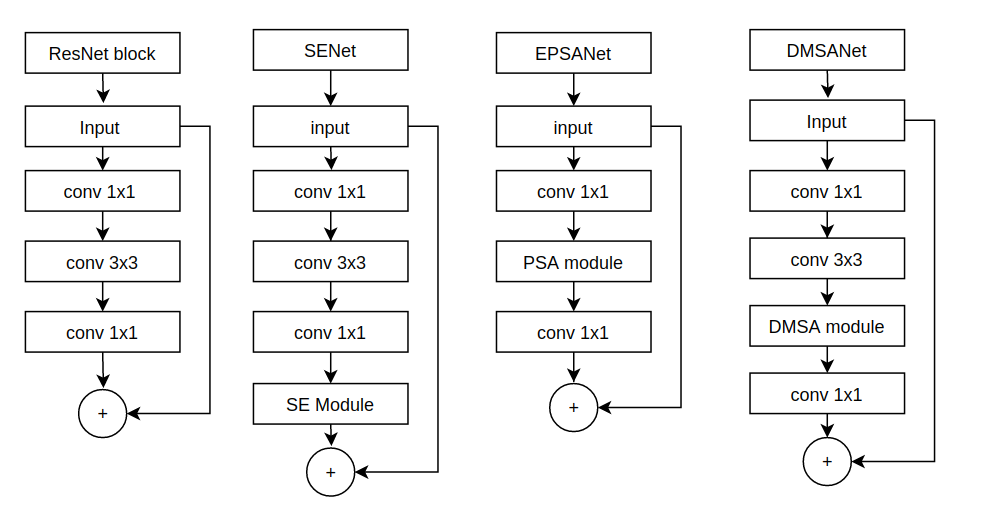}
    \caption{Illustration and comparison of ResNet, SENet, EPSANet and our proposed DMSANet blocks.}
    \label{a1}
\end{figure}

The architectural details our proposed attention network is shown in Table 1:

\begin{table}[hbt!]
  \caption{Network design of the proposed DMSANet.}
  \label{h3}
  \centering
  \begin{tabular}{lll}
  \toprule
Output &ResNet-50 &DMSANet\\
    \midrule

112$\times$112 &7$\times$7, 64 &7$\times$7, 64\\
56$\times$56 &3$\times$3 max pool &3$\times$3 max pool\\
56$\times$56 &

$\begin{bmatrix}
1 $$\times$$ 1, &64\\
3 $$\times$$ 3, &64\\ 
1 $$\times$$ 1, &256\\
\end{bmatrix}$
$\times 3$ &

$\begin{bmatrix}
1 $$\times$$ 1, &64\\
DMSA, &64\\  
1 $$\times$$ 1, &256\\
\end{bmatrix}$ 
$\times$ 3\\

28$\times$28 &

$\begin{bmatrix}
1 $$\times$$ 1, &128\\
3 $$\times$$ 3, &128\\ 
1 $$\times$$ 1, &512\\ 
\end{bmatrix}$ 
$\times$ 4&

$\begin{bmatrix}
1 $$\times$$ 1, &128\\
DMSA, &128\\  
1 $$\times$$ 1, &512\\
\end{bmatrix}$ 
$\times$ 4\\

14$\times$14 &

$\begin{bmatrix}
1 $$\times$$ 1, &256\\
3 $$\times$$ 3, &256\\ 
1 $$\times$$ 1, &1024\\
\end{bmatrix}$ 
$\times$ 6&

$\begin{bmatrix}
1 $$\times$$ 1, &256\\
DMSA, &256\\
1 $$\times$$ 1, &1024\\
\end{bmatrix}$ 
$\times$ 6\\

7$\times$7 &

$\begin{bmatrix}
1 $$\times$$ 1, &512\\
3 $$\times$$ 3, &512\\ 
1 $$\times$$ 1, &2048\\
\end{bmatrix}$ 
$\times$ 3&

$\begin{bmatrix}
1 $$\times$$ 1, &512\\
DMSA, &512\\ 
1 $$\times$$ 1, &2048\\
\end{bmatrix}$ 
$\times$ 3\\

1 $\times$ 1 &7$\times$7 GAP,1000-d fc &7$\times$7 GAP,1000-d fc\\
    \bottomrule
  \end{tabular}
\end{table}

\subsection{Implementation Details}

We use Residual Networks \citep{he2016deep} as the backbone which is widely used in literature for image classification on Imagenet dataset \citep{deng2009imagenet}. Data augmentation is used for increasing the size of the dataset and the input tensor is cropped to size $224\times224$. Stochastic Gradient Descent is used as the optimizer with learning rate of $1e^{-4}$, momentum as 0.9 and mini batch size of 64. The learning rate is initially set as 0.1 and is decreased by a factor of 10 after every 20 epochs for 50 epochs in total.

We use Residual Network along with FPN as the backbone network \citep{lin2017feature} for object detection. The detectors we benchmark against are Faster RCNN \citep{ren2015faster}, Mask RCNN\citep{he2017mask} and RetinaNet \citep{lin2017focal} on MS-COCO dataset \citep{lin2014microsoft}. Stochastic Gradient Descent is used as the optimizer with a weight decay of 1$e^{-4}$, momentum is 0.9, and the batch size is 16 per GPU for 10 epochs. The learning rate is set as 0.01 and is decreased by the factor of 10 every 10th epoch. 

For instance segmentation we use Mask RCNN \citep{he2017mask} with FPN \citep{lin2017feature} as backbone. Stochastic Gradient Descent is used as the optimizer with a weight decay of 1$e^{-4}$, momentum is 0.9, and the batch size is 4 per GPU for 10 epochs. The learning rate is set as 0.01 and is decreased by the factor of 10 every 10th epoch. 

\section{Results}

\subsection{Image Classification on ImageNet}

We compare our network with previous state of the art on ResNet with 50 and 101 layers. 

Our network shows the best performance in accuracy, achieving a considerable improvement compared with all the previous attention models along with lower computational cost. The comparision of our network against previous state of the art with ResNet50 as backbone is shown in Table 2:

\begin{table}[hbt!]
  \caption{Comparison of various attention methods on ImageNet with ResNet50 as backbone in terms of network parameters(in millions), floating point operations per second (FLOPs), Top-1 and Top-5 Validation Accuracy(\%). The best
records are marked in bold.}
  \label{h3}
  \centering
  \begin{tabular}{lllll}
  \toprule
Network &Params &FLOPs &Top-1 Acc (\%) &Top-5 Acc(\%)\\
    \midrule
ResNet &25.56 &4.12G &75.20 &92.52\\
SENet &28.07 &4.13G &76.71 &93.38\\
CBAM &28.07 &4.14G &77.34 &93.69\\
ABN &43.59 &7.18G &76.90 &-\\
GCNet &28.11 &4.13G &77.70 &93.66\\
AANet &25.80 &4.15G &77.70 &93.80\\
ECANet &25.56 &4.13G &77.48 &93.68\\
FcaNet &28.07 &4.13G &78.52 &94.14\\
EPSANet(Small) &\textbf{22.56} &3.62G &77.49 &93.54\\
EPSANet(Large) &27.90 &4.72G &78.64 &94.18\\
DMSANet &26.25 &\textbf{3.44G} &\textbf{80.02} &\textbf{94.27}\\
\bottomrule
  \end{tabular}
\end{table}

The comparision of our network against previous state of the art with ResNet101 as backbone is shown in Table 3:

\begin{table}[hbt!]
  \caption{Comparison of various attention methods on ImageNet with ResNet 101 as backbone in terms of network parameters(in millions), floating point operations per second (FLOPs), Top-1 and Top-5 Validation Accuracy(\%). The best
records are marked in bold.}
  \label{h3}
  \centering
  \begin{tabular}{lllll}
  \toprule
Network &Params &FLOPs &Top-1 Acc (\%) &Top-5 Acc(\%)\\
    \midrule
ResNet &44.55 &7.85G &76.83 &93.48\\
SENet &49.29 &7.86G &77.62 &93.93\\
BAM &44.91 &7.93G &77.56 &93.71\\
CBAM &49.33 &7.88G &78.49 &94.31\\
SRM &44.68 &7.95G &78.47 &94.20\\
ECANet &44.55 &7.86G &78.65 &94.34\\
AANet &45.40 &8.05G &78.70 &94.40\\
EPSANet(Small) &\textbf{38.90} &\textbf{6.82G} &78.43 &94.11\\
EPSANet(Large) &49.59 &8.97G &79.38 &94.58\\
DMSANet &42.29 &7.11G &\textbf{81.54} &\textbf{94.93}\\
    \bottomrule
  \end{tabular}
\end{table}

\subsection{Object Detection on MS COCO}

The comparision of our network using Faster RCNN detector against previous state of the art is shown in Table 4:

\begin{table}[hbt!]
  \caption{Comparison of object detection results on COCO val2017 using Faster RCNN detector. The best
records are marked in bold.}
  \label{h3}
  \centering
  \begin{tabular}{lllllllll}
  \toprule
Backbone &Params(M) &GFLOPs &AP &$AP_{50}$ &$AP_{75}$ &$AP_{S}$ &$AP_{M}$ &$AP_{L}$\\
    \midrule
ResNet-50 &41.53 &207.07 &36.4 &58.2 &39.5 &21.8 &40.0& 46.2\\
SENet-50 &44.02 &207.18 &37.7 &60.1 &40.9 &22.9 &41.9& 48.2\\
ECANet-50 &41.53 &207.18 &38.0 &60.6 &40.9 &23.4 &42.1& 48.0\\
SANet-50&41.53 &207.35 &38.7 &61.2 &41.4 &22.3 &42.5& 49.8\\
FcaNet-50&44.02 &215.63 &39.0 &61.1 &42.3 &23.7 &42.8& 49.6\\
EPSANet-50(Small) &\textbf{38.56} &\textbf{197.07} &39.2 &60.3 &42.3 &22.8 &42.4 &51.1\\
EPSANet-50(Large) &43.85 &219.64 &40.9 &\textbf{62.1} &44.6 &23.6 &44.5 &54.0\\
DMSANet &44.17 &222.31 &\textbf{41.4} &61.9 &\textbf{46.2} &\textbf{25.8} &\textbf{44.7} &\textbf{55.3}\\
 \bottomrule
  \end{tabular}
\end{table}

The comparision of our network using MASK RCNN detector against previous state of the art is shown in Table 5:

\begin{table}[hbt!]
  \caption{Comparison of object detection results on COCO val2017 using Mask RCNN detector. The best
records are marked in bold.}
  \label{h3}
  \centering
  \begin{tabular}{lllllllll}
  \toprule
Backbone &Params(M) &GFLOPs &AP &$AP_{50}$ &$AP_{75}$ &$AP_{S}$ &$AP_{M}$ &$AP_{L}$\\
    \midrule
ResNet-50&44.18 &275.58 &37.2 &58.9 &40.3 &22.2 &40.7 &48.0\\
SENet-50 &46.67 &275.69 &38.7 &60.9 &42.1 &23.4 &42.7 &50.0\\
Non-local &46.50 &288.70 &38.0 &59.8 &41.0 &- &- &-\\
GCNet-50 &46.90 &279.60 &39.4 &61.6 &42.4 &- &- &-\\
ECANet-50 &44.18 &275.69 &39.0 &61.3 &42.1 &24.2 &42.8 &49.9\\
SANet-50&44.18 &275.86 &39.4 &61.5 &42.6 &23.4 &42.8& 51.1\\
FcaNet-50 &46.66 &261.93 &40.3 &62.0 &44.1 &25.2 &43.9& 52.0\\
EPSANet-50(Small) &\textbf{41.20} &\textbf{248.53} &40.0 &60.9 &43.3 &22.3 &43.2 &52.8\\
EPSANet-50(Large)) &46.50 &271.10 &41.4 &\textbf{62.3} &45.3 &23.6& 45.1 &54.6\\
DMSANet &47.23 &279.26 &\textbf{43.1} &61.6 &\textbf{47.5} &\textbf{24.1}& \textbf{46.9} &\textbf{56.5}\\
 \bottomrule
  \end{tabular}
\end{table}

The comparision of our network using RetinaNet detector against previous state of the art is shown in Table 6:

\begin{table}[hbt!]
  \caption{Comparison of object detection results on COCO val2017 using RetinaNet detector. The best
records are marked in bold.}
  \label{h3}
  \centering
  \begin{tabular}{lllllllll}
  \toprule
Backbone &Params(M) &GFLOPs &AP &$AP_{50}$ &$AP_{75}$ &$AP_{S}$ &$AP_{M}$ &$AP_{L}$\\
    \midrule
ResNet-50 &37.74 &239.32 &35.6 &55.5 &38.2 &20.0 &39.6& 46.8\\
SENet-50 &40.25 &239.43 &37.1 &57.2 &39.9 &21.2 &40.7& 49.3\\
SANet-50 &37.74 &239.60 &37.5 &58.5 &39.7 &21.3 &41.2 &45.9\\
EPSANet-50(Small) &\textbf{34.78} &\textbf{229.32} &38.2 &58.1 &40.6 &21.5 &41.5 &51.2\\
EPSANet-50(Large)) &40.07 &251.89 &39.6 &59.4 &42.3 &21.2& 43.4 &52.9\\
DMSANet &41.63 &270.17 &\textbf{40.2} &\textbf{59.8} &\textbf{44.1} &\textbf{23.5}& \textbf{44.8} &\textbf{54.8}\\
    \bottomrule
  \end{tabular}
\end{table}

\subsection{Instance Segmentation on MS COCO}

We used Mask-RCNN \citep{he2017mask} as benchmark on MS-COCO dataset \citep{lin2014microsoft}. The comparision results of our network on instance segmentation using MS COCO dataset against previous state of the art is shown in Table 7:

\begin{table}[hbt!]
  \caption{Instance segmentation results of different attention networks by using the Mask R-CNN on
COCO. The best
records are marked in bold.}
  \label{h3}
  \centering
  \begin{tabular}{lllllll}
  \toprule
Network &AP &$AP_{50}$ &$AP_{75}$ &$AP_{S}$ &$AP_{M}$ &$AP_{L}$\\
    \midrule
ResNet-50 &34.1 &55.5 &36.2 &16.1 &36.7 &50.0\\
SENet-50 &35.4 &57.4 &37.8 &17.1 &38.6 &51.8\\
GCNet &35.7 &58.4 &37.6 &- &- &-\\
ECANet &35.6 &58.1 &37.7 &17.6 &39.0 &51.8\\
FcaNet &36.2 &58.6 &38.1 &- &- &-\\
SANet &36.1 &58.7 &38.2 &19.4 &39.4 &49.0\\
EPSANet-50(Small) &35.9 &57.7 &38.1 &18.5 &38.8 &49.2\\
EPSANet-50(Large) &37.1 &59.0 &39.5 &\textbf{19.6} &40.4 &50.4\\
DMSANet&\textbf{37.4} &\textbf{61.1} &\textbf{40.7} &19.3 &\textbf{40.9} &\textbf{51.7}\\
    \bottomrule
  \end{tabular}
\end{table}

\subsection{Ablation Study}

The ablation studies of our architecture is shown in Table 8.
The results are best obtained using instance normalization. Both removing $F_{c}(·)$ and using $1\times1$ Conv results in reduced performance as compared to the original network. The earlier is because $F_{c}(·)$ is used to enhance the performance of individual features while latter is because number of channels in each sub-feature is too few, so it is not important to exchange information among different channels. 

\begin{table}[hbt!]
  \caption{Performance comparisons of our network using ResNet 50 as backbone with four options (i.e., original, using Batch Normalization, using Group Normalization, using shuffle normalization, eliminating $F_{c}(·)$ and using $1\times1$ Conv to replace $F_{c}(·)$ on ImageNet-1k in terms of GFLOPs and Top-1/Top-5 accuracy (in \%). The best
records are marked in bold.}
  \label{h3}
  \centering
  \begin{tabular}{llll}
  \toprule
Methods &GFLOPs &Top-1 Acc(\%) &Top-5 Acc(\%)\\
    \midrule
origin & \textbf{3.44} &\textbf{80.02} &\textbf{94.27}\\
W BN &3.82 &77.37 &93.80\\
W GN &3.56 &77.61 &92.89\\
W SN &3.51 &78.16 &93.48\\
W/O $F_{c}(·)$ &4.07 &77.64 &93.18\\
$1\times1$ Conv &3.55 &78.69 &93.71\\
    \bottomrule
  \end{tabular}
\end{table}

\section{Conclusions}

In this paper, we propose a novel Attention module named Dual Multi Scale Attention Network (DMSANet). Our network is comprised of two parts 1) first for aggregating feature information at various scales 2) second made up of position and channel attention modules in parallel for capturing global contextual information. After evaluating our network both qualitatively and quantitatively, we show that our network outperforms previous state of the art across image classification, object detection and instance
segmentation problems. The ablation experiments show that our attention module captures long-range contextual information effectively at various scales thus making it generalizable to other tasks. The best part of DMSANet attention module is that it is very lightweight and hence could be easily plugged into various custom networks as and when required.

\subsubsection*{Acknowledgments}

We would like to thank Nvidia for providing the GPUs for this work.

\bibliography{neurips_2019}

\begin{thebibliography}{32}
\providecommand{\natexlab}[1]{#1}
\providecommand{\url}[1]{\texttt{#1}}
\expandafter\ifx\csname urlstyle\endcsname\relax
  \providecommand{\doi}[1]{doi: #1}\else
  \providecommand{\doi}{doi: \begingroup \urlstyle{rm}\Url}\fi

\bibitem[Bello et~al.(2019)Bello, Zoph, Vaswani, Shlens, and
  Le]{bello2019attention}
I.~Bello, B.~Zoph, A.~Vaswani, J.~Shlens, and Q.~V. Le.
\newblock Attention augmented convolutional networks.
\newblock In \emph{Proceedings of the IEEE/CVF International Conference on
  Computer Vision}, pages 3286--3295, 2019.

\bibitem[Cao et~al.(2019)Cao, Xu, Lin, Wei, and Hu]{cao2019gcnet}
Y.~Cao, J.~Xu, S.~Lin, F.~Wei, and H.~Hu.
\newblock Gcnet: Non-local networks meet squeeze-excitation networks and
  beyond.
\newblock In \emph{Proceedings of the IEEE/CVF International Conference on
  Computer Vision Workshops}, pages 0--0, 2019.

\bibitem[Chen et~al.(2018)Chen, Zhu, Papandreou, Schroff, and
  Adam]{chen2018encoder}
L.-C. Chen, Y.~Zhu, G.~Papandreou, F.~Schroff, and H.~Adam.
\newblock Encoder-decoder with atrous separable convolution for semantic image
  segmentation.
\newblock In \emph{Proceedings of the European conference on computer vision
  (ECCV)}, pages 801--818, 2018.

\bibitem[Deng et~al.(2009)Deng, Dong, Socher, Li, Li, and
  Fei-Fei]{deng2009imagenet}
J.~Deng, W.~Dong, R.~Socher, L.-J. Li, K.~Li, and L.~Fei-Fei.
\newblock Imagenet: A large-scale hierarchical image database.
\newblock In \emph{2009 IEEE conference on computer vision and pattern
  recognition}, pages 248--255. Ieee, 2009.

\bibitem[Duta et~al.(2020)Duta, Liu, Zhu, and Shao]{duta2020pyramidal}
I.~C. Duta, L.~Liu, F.~Zhu, and L.~Shao.
\newblock Pyramidal convolution: rethinking convolutional neural networks for
  visual recognition.
\newblock \emph{arXiv preprint arXiv:2006.11538}, 2020.

\bibitem[Fu et~al.(2019)Fu, Liu, Tian, Li, Bao, Fang, and Lu]{fu2019dual}
J.~Fu, J.~Liu, H.~Tian, Y.~Li, Y.~Bao, Z.~Fang, and H.~Lu.
\newblock Dual attention network for scene segmentation.
\newblock In \emph{Proceedings of the IEEE/CVF Conference on Computer Vision
  and Pattern Recognition}, pages 3146--3154, 2019.

\bibitem[Fukui et~al.(2019)Fukui, Hirakawa, Yamashita, and
  Fujiyoshi]{fukui2019attention}
H.~Fukui, T.~Hirakawa, T.~Yamashita, and H.~Fujiyoshi.
\newblock Attention branch network: Learning of attention mechanism for visual
  explanation.
\newblock In \emph{Proceedings of the IEEE/CVF Conference on Computer Vision
  and Pattern Recognition}, pages 10705--10714, 2019.

\bibitem[Gao et~al.(2018)Gao, Wang, and Ji]{gao2018channelnets}
H.~Gao, Z.~Wang, and S.~Ji.
\newblock Channelnets: Compact and efficient convolutional neural networks via
  channel-wise convolutions.
\newblock \emph{arXiv preprint arXiv:1809.01330}, 2018.

\bibitem[Gao et~al.(2019)Gao, Cheng, Zhao, Zhang, Yang, and
  Torr]{gao2019res2net}
S.~Gao, M.-M. Cheng, K.~Zhao, X.-Y. Zhang, M.-H. Yang, and P.~H. Torr.
\newblock Res2net: A new multi-scale backbone architecture.
\newblock \emph{IEEE transactions on pattern analysis and machine
  intelligence}, 2019.

\bibitem[He et~al.(2016)He, Zhang, Ren, and Sun]{he2016deep}
K.~He, X.~Zhang, S.~Ren, and J.~Sun.
\newblock Deep residual learning for image recognition.
\newblock In \emph{Proceedings of the IEEE conference on computer vision and
  pattern recognition}, pages 770--778, 2016.

\bibitem[He et~al.(2017)He, Gkioxari, Doll{\'a}r, and Girshick]{he2017mask}
K.~He, G.~Gkioxari, P.~Doll{\'a}r, and R.~Girshick.
\newblock Mask r-cnn.
\newblock In \emph{Proceedings of the IEEE international conference on computer
  vision}, pages 2961--2969, 2017.

\bibitem[Hu et~al.(2018{\natexlab{a}})Hu, Shen, Albanie, Sun, and
  Vedaldi]{hu2018gather}
J.~Hu, L.~Shen, S.~Albanie, G.~Sun, and A.~Vedaldi.
\newblock Gather-excite: Exploiting feature context in convolutional neural
  networks.
\newblock \emph{arXiv preprint arXiv:1810.12348}, 2018{\natexlab{a}}.

\bibitem[Hu et~al.(2018{\natexlab{b}})Hu, Shen, and Sun]{hu2018squeeze}
J.~Hu, L.~Shen, and G.~Sun.
\newblock Squeeze-and-excitation networks.
\newblock In \emph{Proceedings of the IEEE conference on computer vision and
  pattern recognition}, pages 7132--7141, 2018{\natexlab{b}}.

\bibitem[Li et~al.(2019)Li, Wang, Hu, and Yang]{li2019selective}
X.~Li, W.~Wang, X.~Hu, and J.~Yang.
\newblock Selective kernel networks.
\newblock In \emph{Proceedings of the IEEE/CVF Conference on Computer Vision
  and Pattern Recognition}, pages 510--519, 2019.

\bibitem[Lin et~al.(2014)Lin, Maire, Belongie, Hays, Perona, Ramanan,
  Doll{\'a}r, and Zitnick]{lin2014microsoft}
T.-Y. Lin, M.~Maire, S.~Belongie, J.~Hays, P.~Perona, D.~Ramanan,
  P.~Doll{\'a}r, and C.~L. Zitnick.
\newblock Microsoft coco: Common objects in context.
\newblock In \emph{European conference on computer vision}, pages 740--755.
  Springer, 2014.

\bibitem[Lin et~al.(2017{\natexlab{a}})Lin, Doll{\'a}r, Girshick, He,
  Hariharan, and Belongie]{lin2017feature}
T.-Y. Lin, P.~Doll{\'a}r, R.~Girshick, K.~He, B.~Hariharan, and S.~Belongie.
\newblock Feature pyramid networks for object detection.
\newblock In \emph{Proceedings of the IEEE conference on computer vision and
  pattern recognition}, pages 2117--2125, 2017{\natexlab{a}}.

\bibitem[Lin et~al.(2017{\natexlab{b}})Lin, Goyal, Girshick, He, and
  Doll{\'a}r]{lin2017focal}
T.-Y. Lin, P.~Goyal, R.~Girshick, K.~He, and P.~Doll{\'a}r.
\newblock Focal loss for dense object detection.
\newblock In \emph{Proceedings of the IEEE international conference on computer
  vision}, pages 2980--2988, 2017{\natexlab{b}}.

\bibitem[Ren et~al.(2015)Ren, He, Girshick, and Sun]{ren2015faster}
S.~Ren, K.~He, R.~Girshick, and J.~Sun.
\newblock Faster r-cnn: Towards real-time object detection with region proposal
  networks.
\newblock \emph{arXiv preprint arXiv:1506.01497}, 2015.

\bibitem[Sagar(2021)]{sagar2021aa3dnet}
A.~Sagar.
\newblock Aa3dnet: Attention augmented real time 3d object detection.
\newblock \emph{arXiv preprint arXiv:2107.12137}, 2021.

\bibitem[Sagar and Soundrapandiyan(2020)]{sagar2020semantic}
A.~Sagar and R.~Soundrapandiyan.
\newblock Semantic segmentation with multi scale spatial attention for self
  driving cars.
\newblock \emph{arXiv preprint arXiv:2007.12685}, 2020.

\bibitem[Sang et~al.(2020)Sang, Zhou, and Zhao]{sang2020pcanet}
H.~Sang, Q.~Zhou, and Y.~Zhao.
\newblock Pcanet: Pyramid convolutional attention network for semantic
  segmentation.
\newblock \emph{Image and Vision Computing}, 103:\penalty0 103997, 2020.

\bibitem[Szegedy et~al.(2016)Szegedy, Vanhoucke, Ioffe, Shlens, and
  Wojna]{szegedy2016rethinking}
C.~Szegedy, V.~Vanhoucke, S.~Ioffe, J.~Shlens, and Z.~Wojna.
\newblock Rethinking the inception architecture for computer vision.
\newblock In \emph{Proceedings of the IEEE conference on computer vision and
  pattern recognition}, pages 2818--2826, 2016.

\bibitem[Wang et~al.(2017)Wang, Jiang, Qian, Yang, Li, Zhang, Wang, and
  Tang]{wang2017residual}
F.~Wang, M.~Jiang, C.~Qian, S.~Yang, C.~Li, H.~Zhang, X.~Wang, and X.~Tang.
\newblock Residual attention network for image classification.
\newblock In \emph{Proceedings of the IEEE conference on computer vision and
  pattern recognition}, pages 3156--3164, 2017.

\bibitem[Wang et~al.(2020)Wang, Wu, Zhu, Li, Zuo, and Hu]{wang2020ecanet}
Q.~Wang, B.~Wu, P.~Zhu, P.~Li, W.~Zuo, and Q.~Hu.
\newblock Eca-net: Efficient channel attention for deep convolutional neural
  networks, 2020.

\bibitem[Wang et~al.(2018)Wang, Girshick, Gupta, and He]{wang2018non}
X.~Wang, R.~Girshick, A.~Gupta, and K.~He.
\newblock Non-local neural networks.
\newblock In \emph{Proceedings of the IEEE conference on computer vision and
  pattern recognition}, pages 7794--7803, 2018.

\bibitem[Woo et~al.(2018)Woo, Park, Lee, and Kweon]{woo2018cbam}
S.~Woo, J.~Park, J.-Y. Lee, and I.~S. Kweon.
\newblock Cbam: Convolutional block attention module.
\newblock In \emph{Proceedings of the European conference on computer vision
  (ECCV)}, pages 3--19, 2018.

\bibitem[Wu and He(2018)]{wu2018group}
Y.~Wu and K.~He.
\newblock Group normalization.
\newblock In \emph{Proceedings of the European conference on computer vision
  (ECCV)}, pages 3--19, 2018.

\bibitem[Zhang et~al.(2020)Zhang, Wu, Zhang, Zhu, Lin, Zhang, Sun, He, Mueller,
  Manmatha, et~al.]{zhang2020resnest}
H.~Zhang, C.~Wu, Z.~Zhang, Y.~Zhu, H.~Lin, Z.~Zhang, Y.~Sun, T.~He, J.~Mueller,
  R.~Manmatha, et~al.
\newblock Resnest: Split-attention networks.
\newblock \emph{arXiv preprint arXiv:2004.08955}, 2020.

\bibitem[Zhang et~al.(2021)Zhang, Zu, Lu, Zou, and Meng]{zhang2021epsanet}
H.~Zhang, K.~Zu, J.~Lu, Y.~Zou, and D.~Meng.
\newblock Epsanet: An efficient pyramid split attention block on convolutional
  neural network.
\newblock \emph{arXiv preprint arXiv:2105.14447}, 2021.

\bibitem[Zhang and Yang(2021)]{zhang2021sa}
Q.-L. Zhang and Y.-B. Yang.
\newblock Sa-net: Shuffle attention for deep convolutional neural networks.
\newblock In \emph{ICASSP 2021-2021 IEEE International Conference on Acoustics,
  Speech and Signal Processing (ICASSP)}, pages 2235--2239. IEEE, 2021.

\bibitem[Zhao et~al.(2017)Zhao, Shi, Qi, Wang, and Jia]{zhao2017pyramid}
H.~Zhao, J.~Shi, X.~Qi, X.~Wang, and J.~Jia.
\newblock Pyramid scene parsing network.
\newblock In \emph{Proceedings of the IEEE conference on computer vision and
  pattern recognition}, pages 2881--2890, 2017.

\bibitem[Zhong et~al.(2020)Zhong, Lin, Bidart, Hu, Daya, Li, Zheng, Li, and
  Wong]{zhong2020squeeze}
Z.~Zhong, Z.~Q. Lin, R.~Bidart, X.~Hu, I.~B. Daya, Z.~Li, W.-S. Zheng, J.~Li,
  and A.~Wong.
\newblock Squeeze-and-attention networks for semantic segmentation.
\newblock In \emph{Proceedings of the IEEE/CVF Conference on Computer Vision
  and Pattern Recognition}, pages 13065--13074, 2020.

\end{thebibliography}

\end{document}